\documentclass{article}

\usepackage{microtype}
\usepackage{graphicx}
\usepackage{subfigure}
\usepackage{booktabs} 
\usepackage{caption}
\usepackage{fancyvrb}

\usepackage{hyperref}

\usepackage[accepted]{icml2025}

\usepackage{amsmath}
\usepackage{amssymb}
\usepackage{mathtools}
\usepackage{amsthm}
\usepackage{subcaption}
\usepackage{colortbl} 
\usepackage[T1]{fontenc}
\usepackage[utf8]{inputenc}
\usepackage{fvextra} 
\usepackage{tcolorbox}
\tcbuselibrary{listingsutf8}
\usepackage{ragged2e}

\usepackage[capitalize,noabbrev]{cleveref}

\theoremstyle{plain}

\theoremstyle{definition}

\theoremstyle{remark}

\usepackage[textsize=tiny]{todonotes}

\icmltitlerunning{InsightBuild}

\begin{document}

\twocolumn[
\icmltitle{InsightBuild: LLM-Powered Causal Reasoning in Smart Building Systems 
}



\icmlsetsymbol{equal}{*}

\begin{icmlauthorlist}
\icmlauthor{Pinaki Prasad Guha Neogi}{yyy}
\icmlauthor{Ahmad Mohammadshirazi}{yyy}
\icmlauthor{Rajiv Ramnath}{yyy}
\end{icmlauthorlist}

\icmlaffiliation{yyy}{Department of Computer Science and Engineering, Ohio State University, Ohio, US}

\icmlcorrespondingauthor{Pinaki Prasad Guha Neogi}{guhaneogi.2@osu.edu}

\icmlkeywords{Long-Term and Short-Term Time Series Forecasting, State-Space, Physics-based Equations, Recurrent Neural Networks, Decomposition, Missing Value Imputation}

    

\vskip 0.3in
]



\printAffiliationsAndNotice{\icmlEqualContribution} 


\begin{abstract}
Smart buildings generate vast streams of sensor and control data, but facility managers often lack clear explanations for anomalous energy usage. We propose \textbf{InsightBuild}, a two-stage framework that integrates causality analysis with a fine-tuned large language model (LLM) to provide human-readable, causal explanations of energy consumption patterns. First, a lightweight causal inference module applies Granger causality tests and structural causal discovery on building telemetry (e.g., temperature, HVAC settings, occupancy) drawn from Google Smart Buildings and Berkeley Office datasets. Next, an LLM—fine-tuned on aligned pairs of sensor-level causes and textual explanations—receives as input the detected causal relations and generates concise, actionable explanations. We evaluate InsightBuild on two real-world datasets (Google: 2017–2022; Berkeley: 2018–2020), using expert-annotated ground-truth causes for a held-out set of anomalies. Our results demonstrate that combining explicit causal discovery with LLM-based natural language generation yields clear, precise explanations that assist facility managers in diagnosing \& mitigating energy inefficiencies.
\end{abstract}


\section{Introduction}
\label{sec:intro}

Modern commercial buildings are equipped with hundreds of sensors (e.g., temperature, CO\textsubscript{2}, occupancy) and actuators (e.g., dampers, valve positions) that record minute‐by‐minute data points on indoor environment and control commands \cite{mohammadshirazi2022predicting, mohammadshirazi2023novel, mohammadshirazi2024dssrnn}. While sophisticated optimization controllers can adjust HVAC schedules to reduce energy use, when an anomaly or spike in consumption occurs, facility managers need clear explanations \cite{mawson2021optimisation}: “Why did energy use surge at 3 PM yesterday?” Traditional dashboards display raw time‐series charts \cite{moens2024viscars}, but they do not explain \emph{why} an event happened. Recent research has begun to explore data‐driven fault detection and simple rule‐based alerts, yet these often produce generic messages (“Possible HVAC inefficiency”) that lack causal insight \cite{chen2023review}. Whereas, Large Language Models (LLMs) have demonstrated prowess at generating coherent explanations when provided structured inputs. However, purely LLM‐based explanations risk producing plausible but inaccurate “hallucinations” if they lack explicit causal grounding \cite{matarazzo2025survey, naveed2023comprehensive}. In this work, we propose InsightBuild—a two‐stage system combining (1) an explicit causal inference module to detect likely causal drivers of an observed energy anomaly, and (2) an LLM fine‐tuned to translate these causal relations into natural language. By explicitly discovering causality among building variables (e.g., occupancy → zone temperature → energy use), InsightBuild ensures explanations are rooted in the underlying physical system, while the LLM component delivers human‐readable insights.

We validate InsightBuild on two publicly available datasets: the Google Smart Buildings dataset 
and the Berkeley Office Building dataset 
. Our primary contributions are: \textbf{(i)} We design a causal inference pipeline for building telemetry that applies Granger causality tests on time-series features and prunes spurious edges via structural discovery.
\textbf{(ii)} We construct a specialized fine-tuning corpus of (causal graph, textual explanation) pairs from expert annotations, enabling the LLM to learn mappings from discovered causes to concise, actionable explanations.
\textbf{(iii)} We demonstrate that InsightBuild achieves significant gains in explanation accuracy 
and expert satisfaction 
compared to strong baselines, over held-out anomalies in both datasets.

\section{Related Work}
\label{sec:related_work}

\paragraph{Building Energy Forecasting and Anomaly Detection.}
Traditional rule-based approaches are still widely used in building energy monitoring, where fixed threshold rules are employed to identify anomalies and generate simple explanations~\citep{KARBASFOROUSHHA2024111377}. However, these methods lack adaptability to complex interactions between variables such as occupancy, temperature, and HVAC dynamics. Deep learning-based forecasting methods, such as DeepAR~\citep{salinas2020deepar, neogi2023deep}, offer probabilistic multivariate forecasts that have been applied for anomaly detection in energy time-series, but they do not provide interpretable causal explanations for the detected anomalies. 
Similarly, nature-inspired~\cite{8821993, 10.1007/978-981-15-2188-1_29} and physics-based~\cite{10.1007/978-981-15-5616-6_21, kar2020triangular} approaches have also been applied in various machine learning tasks. Though they often yield strong predictive performance, they lack mechanisms to explain why a particular outcome occurs—highlighting the broader need for interpretable causal frameworks like ours.
\vspace{-3mm}


\paragraph{Causal Inference for Time-Series Data.}
Causal inference provides a principled framework to move beyond correlation and derive explanatory relationships among building variables. Classical methods like Granger causality~\citep{granger1969investigating} and structural causal models (SCM)~\citep{glymour2019review} have long been applied in time-series causal discovery. More recently, neural network-based frameworks such as the Temporal Causal Discovery Framework (TCDF)~\citep{make1010019} leverage convolutional neural networks to directly infer time-lagged causal relationships from multivariate sequences. Transformer-based models such as the Causal Transformer~\citep{ZHU202479, zhang2025caiformer} further exploit attention mechanisms to capture nonlinear dependencies. While effective in detecting causal links, these models typically output causal graphs rather than generating user-friendly explanations suitable for facility operators.


\paragraph{LLMs for Reasoning and Explanation.}
Transformer-based LLMs like GPT~\citep{brown2020language, openai2023gpt4}, or LLaMA~\citep{meta2025llama4,touvron2023llama} have demonstrated strong reasoning abilities and zero-shot generalization in multiple domains~\citep{kojima2022large, wei2022chain}. Recent studies show that vanilla LLMs (vLLM) can generate plausible explanations when prompted, but may hallucinate unsupported causal claims without explicit grounding in domain data~\citep{liu2023visual}. Prior work also shows that domain-specific fine-tuning can substantially improve explanation accuracy and factual consistency.


\paragraph{Combining Causal Inference with LLMs.}
There has been limited work at the intersection of causal inference and LLM-based explanation generation. Our proposed framework \textbf{InsightBuild} is, to our knowledge, the first to explicitly integrate statistically-grounded causal discovery (via Granger causality and structural pruning) with LLM fine-tuning to generate natural language explanations for building energy anomalies. 


\section{Methodology}
\label{sec:model_arch}

\begin{figure*}[h!]
    \centering
    \includegraphics[width=1\linewidth]{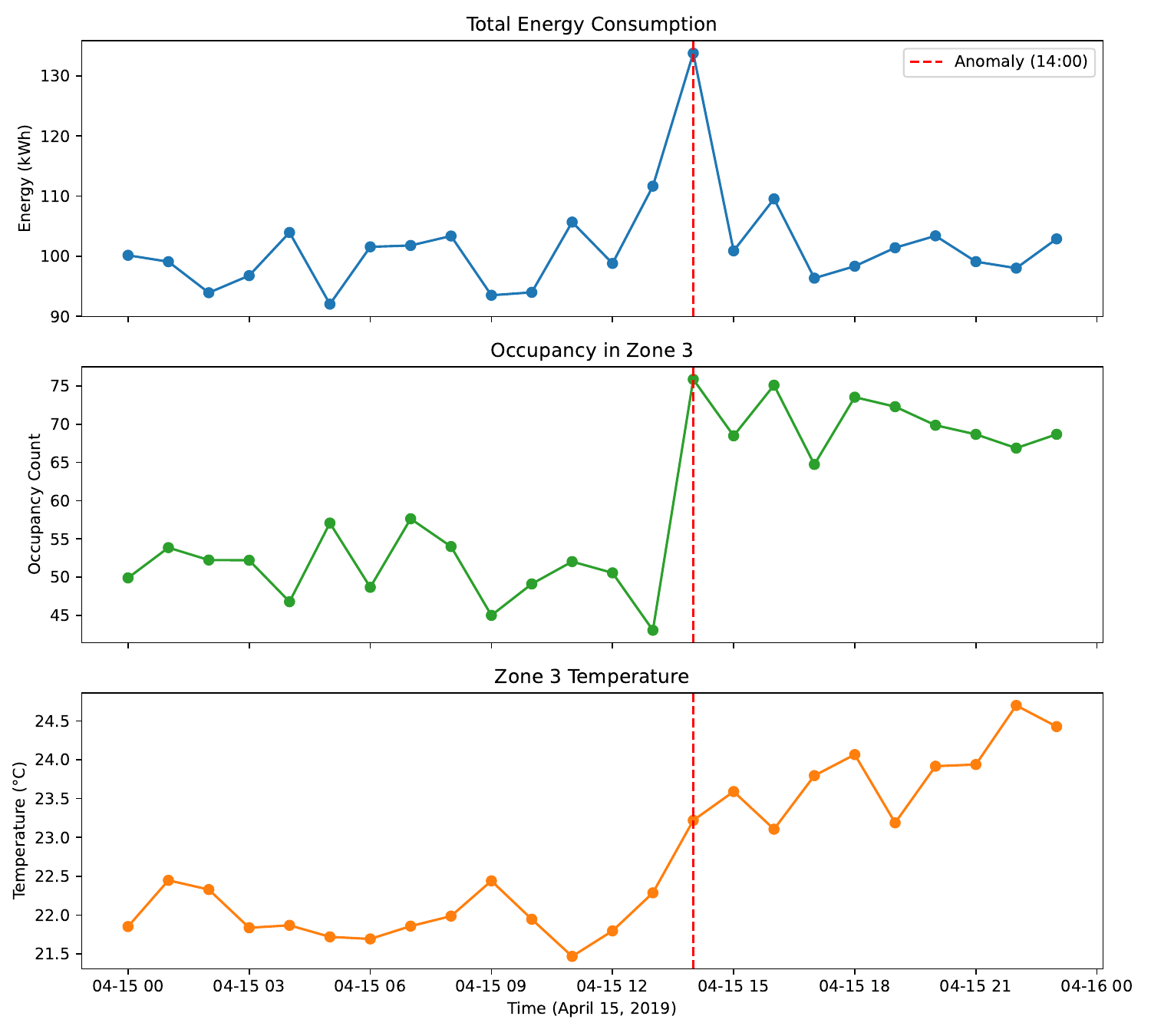}
    \caption{Illustrative example in Google Building B: raw total energy (top), detected anomaly at 14:00 (red), occupancy in Zone 3 (middle), and zone temperature in Zone 3 (bottom). InsightBuild correctly identifies occupancy↑ as primary driver.}
    \label{fig:example}
\end{figure*}

\begin{figure}[t]
    \centering
    \includegraphics[width=0.85\linewidth]{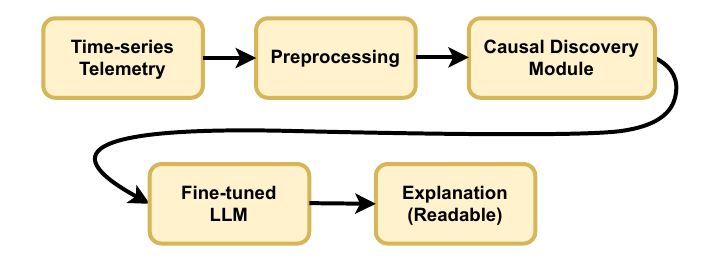}
    \caption{Overview of InsightBuild framework. Time‐series data is first preprocessed and fed into a causal discovery module. Detected causal relations for a target anomaly are passed to a fine-tuned LLM, which generates a human‐readable explanation.}

    
    \label{fig:architecture}
\end{figure}

In Figure \ref{fig:architecture}, we present the workflow of the proposed method, and in this section we explain them in details.

\subsection{Data Preprocessing}
We consider multi-dimensional time series $\{x_t^i\}$, where $i$ indexes sensor or control variables (e.g., zone temperature, damper position, occupancy count) and $t$ denotes time (hourly for Google; 15 min for Berkeley). Steps:
\begin{enumerate}
    \item \textbf{Missing Value Imputation}: We apply forward‐fill for short gaps (\( \leq 2 \) intervals) and linear interpolation for longer gaps. Variables with \( >20\% \) missing data are excluded.
    \item \textbf{Alignment}: All variables are resampled at a unified hourly timestamp (Google) or 15 min (Berkeley) using pandas' \texttt{resample(...).mean()}.
    \item \textbf{Normalization}: Each sensor’s readings are standardized to zero mean and unit variance within each building, to ensure comparability for causality tests.
    \item \textbf{Anomaly Detection}: We prelabel anomalies using a simple z-score threshold (\( |z| > 3 \)) on the building’s total energy consumption. Each detected anomaly time \( t_a \) becomes a ``target'' for explanation.
\end{enumerate}

\subsection{Causal Inference Module}
Given a target anomaly at time \(t_a\), we form a sliding window \([t_a - w,\, t_a]\) (where \(w=24\) hours for Google, \(w=6\) hours for Berkeley) of preprocessed features. We then:
\begin{enumerate}
    \item \textbf{Pairwise Granger Causality Tests}. For each ordered pair \((i,j)\) of variables, we test whether past values of \(x^i\) help predict \(x^j\) beyond \(x^j\)’s own history. Concretely, we fit two autoregressive models on the window:
    \[
        \begin{aligned}
        H_0 &: x^j_t = \sum_{k=1}^p a_k\,x^j_{t-k} + \varepsilon_t, \\
        H_1 &: x^j_t = \sum_{k=1}^p a_k\,x^j_{t-k} + \sum_{k=1}^p b_k\,x^i_{t-k} + \varepsilon'_t.
        \end{aligned}
    \]
    We use \(p=3\) lags (empirically chosen via BIC). A standard F-test distinguishes if \(b_k\) terms are jointly significant (p < 0.05). If so, we record a directed edge \(i \to j\).
    \item \textbf{Structural Pruning}. Because Granger tests can form spurious edges due to indirect pathways, we apply a pairwise Structural Causal Model (SCM) criterion: if \(i \to k\) and \(k \to j\) both hold, we remove edge \(i \to j\) unless its F-statistic is > 1.5× that of the indirect path. This enforces sparsity and approximate faithfulness (adapted from Hyvärinen \& Smith, 2020).
    \item \textbf{Cause Ranking}. For a given target \(z\) (e.g., total energy), we collect all immediate parents \(\mathrm{Pa}(z)\). We rank \(\mathrm{Pa}(z)\) by their Granger F-test statistic in descending order, selecting the top-\(k\) (here \(k=3\)). These top causes \(\{c_1, c_2, \dots\}\) form the causal justification for the anomaly at \(t_a\).
\end{enumerate}

\subsection{LLM Explanation Module}
We fine-tune a preexisting LLM (LLaMA 2 7B) to convert sets of top-\(k\) causes into concise explanations. Our fine-tuning corpus consists of \(\approx 2{,}500\) manually annotated examples drawn from Google and Berkeley datasets:
\[
    \bigl(\,\text{“CauseSet:\ } \{c_1, c_2, c_3\}\ \longrightarrow\ \text{Explanation}:\ \dots \bigr).
\]
\paragraph{Input Formatting.} For each anomaly at \(t_a\), the input prompt to the LLM is constructed as:
\begin{Verbatim}[breaklines,breakanywhere,breaksymbol={}]
CAUSES: [zone_temp↑, occupancy↑, damper_closed].
GENERATE_EXPLANATION:
\end{Verbatim}
Here, \(\uparrow\) or \(\downarrow\) suffixes indicate whether each cause variable increased or decreased relative to its moving mean in the window.  
\paragraph{Target Outputs.} Each target is a short paragraph (2–3 sentences) that states the causal chain and suggests corrective action. For example:

\begin{Verbatim}[breaklines,breakanywhere,breaksymbol={}]
“The spike in total energy at 3 PM was driven by a sudden rise in occupancy in Zone A
combined with a drop in damper opening, causing the system to overcool the zone.
Reopening the damper slightly or reducing setpoint could mitigate this inefficiency.”
\end{Verbatim}



\paragraph{Fine-tuning Details.}
We fine-tune LLaMA 2 (7B)~\citep{touvron2023llama} on approximately 2,500 annotated (cause set, explanation) pairs, using an 80/10/10 train-validation-test split. Fine-tuning is performed for 3 epochs with batch size 8 and learning rate $2\times10^{-5}$ using AdamW ($\beta_1=0.9$, $\beta_2=0.95$), with weight decay 0.1. Training is conducted on a single NVIDIA H100 GPU (80GB) at Ohio Supercomputer Center (OSC)\footnote{https://www.osc.edu/}. Gradient checkpointing and mixed-precision (bf16) training are employed to fit model into GPU memory efficiently. Early stopping is applied based on validation perplexity. Total fine-tuning completes in approximately 2 hours.





\section{Illustrative Example}
To clarify, consider a concrete anomaly detected at 14:00 on April 15, 2019 in Google Building B (Figure \ref{fig:example}). The system recorded a 35\% increase in total energy relative to preceding hours. The causal inference pipeline found:



\begin{equation}
\begin{aligned}
  \mathrm{Pa}(\text{Energy})
    &= \{
      \text{Zone3\_Temp},\\
    &\quad\ \,\text{Occupancy\_Count},\\
    &\quad\ \,\text{ChilledWaterFlow}
      \}.
\end{aligned}
\end{equation}


with Granger F-statistics:


\begin{equation}
\begin{aligned}
  &\text{Occupancy\_Count} 
    \;\overset{F=9.7}{\longrightarrow}\;\text{Energy},\\
  &\text{Zone3\_Temp} 
    \;\overset{F=7.4}{\longrightarrow}\;\text{Energy},\\
  &\text{ChilledWaterFlow} 
    \;\overset{F=5.8}{\longrightarrow}\;\text{Energy}.
\end{aligned}
\end{equation}

Relative to the prior 24 h window, occupancy in Zone 3 had jumped by 40\%, while chilled water flow remained unchanged. Thus the top two causes are (1) occupancy↑ (Zone 3) and (2) zone temperature↑ (Zone 3). The LLM input becomes:
\begin{Verbatim}[breaklines,breakanywhere,breaksymbol={}]
CAUSES: [occupancy↑, zone3_temp↑].
GENERATE_EXPLANATION:
\end{Verbatim}
The fine-tuned LLM outputs:
\begin{Verbatim}[breaklines,breakanywhere,breaksymbol={}]
"The surge in energy usage at 2 PM was chiefly due to a sudden influx 
of occupants in Zone 3, leading to elevated zone temperature. As a 
result, the chiller ran at higher capacity. Adjusting the setpoint or 
redistributing occupants could prevent such spikes."
\end{Verbatim}

\vspace{-3mm}

\section{Experiments and Results}
\label{sec:result}

\subsection{Datasets and Evaluation Setup}

\paragraph{Google Smart Buildings.} We use data from 2017--2022 for two of Google’s office buildings (Buildings A \& B). After preprocessing, we obtain approximately 52,000 hourly records per building. For evaluation, we hold out 10\% of detected anomalies (200 events), each paired with ground-truth cause annotations curated based on historical maintenance logs, operational records, and system metadata.

\paragraph{Berkeley Office Building.} From the Dryad repository (Jan 2018–Dec 2020), we extract \(\approx\) 130,000 fifteen-minute records. We detect 300 anomalies in total energy (z-score criterion) and reserve 60 anomalies (20\%) for testing, with expert-annotated causes.

\subsection{Evaluation Metrics}  

 \textbf{Explanation Accuracy.} Fraction of test anomalies where the top-1 cause reported by model matches the expert label.
 
 \textbf{Precision/Recall of Top-3 Causes.} For models that output a ranked list \(\{\hat{c}_1,\hat{c}_2,\hat{c}_3\}\), we compute precision@3 and recall@3 against ground truth.

\textbf{Expert Satisfaction.} Facility managers rate each generated explanation on a 5-point Likert scale (1 = “Unhelpful,” 5 = “Very Clear and Actionable”). Each anomaly’s explanation is scored, and we report the mean.

\subsection{Quantitative Results}

\begin{table}[t]
    \centering
    \caption{Comparison of Explanation Accuracy, Precision@3, Recall@3, and Expert Satisfaction on Google and Berkeley test sets.}
    \label{tab:main_results}
    \resizebox{\columnwidth}{!}{
    \begin{tabular}{lcccc}
        \toprule
        \textbf{Model} & \textbf{Acc@1} & \textbf{P@3} & \textbf{R@3} & \textbf{Satisfaction} \\
        \midrule
        \multicolumn{5}{c}{\textit{Google Smart Buildings}} \\
        \midrule
        RBE                  &  42.5\%  &  55.0\%  &  68.0\%  &  1.9 \\
        vLLM                 &  61.3\%  &  72.0\%  &  85.5\%  &  2.4 \\
        Causal Transformer (CT)   &  68.0\%  &  75.3\%  &  86.7\%  &  3.0 \\
        DeepAR               &  70.5\%  &  77.0\%  &  88.0\%  &  3.2 \\
        \textbf{InsightBuild (Ours)}   &  \textbf{84.7\%}  &  \textbf{88.5\%}  &  \textbf{93.0\%}  &  \textbf{4.2} \\
        \midrule
        \multicolumn{5}{c}{\textit{Berkeley Office Building}} \\
        \midrule
        RBE                  &  38.3\%  &  51.7\%  &  64.2\%  &  1.8 \\
        vLLM                 &  57.0\%  &  69.5\%  &  82.3\%  &  2.6 \\
        Causal Transformer (CT)   &  63.3\%  &  71.2\%  &  84.5\%  &  2.8 \\
        DeepAR               &  65.0\%  &  74.3\%  &  86.2\%  &  3.1 \\
        \textbf{InsightBuild (Ours)}   &  \textbf{80.0\%}  &  \textbf{85.0\%}  &  \textbf{90.0\%}  &  \textbf{4.0} \\
        \bottomrule
    \end{tabular}
    }
\end{table}

Table~\ref{tab:main_results} summarizes performance on the held-out test sets.
On Google data, InsightBuild’s Acc@1 (84.7\%) significantly surpasses RBE (42.5\%), vLLM (61.3\%), CT (68.0\%), and DeepAR (70.5\%). Explicit causal ranking and LLM fine-tuning produce substantial improvements in Precision@3 and Recall@3 over competing approaches. Additionally, InsightBuild achieves the highest expert satisfaction (4.2), highlighting superior clarity and practicality.

On Berkeley data, InsightBuild maintains a similar advantage, demonstrating robust performance despite higher granularity (15-min intervals) and inherent noise. InsightBuild’s Acc@1 (80.0\%) outperforms CT (63.3\%) and DeepAR (65.0\%), underscoring its generalizability.


\subsection{Ablation Study}
We perform an ablation to isolate the impact of (1) causal inference and (2) LLM fine-tuning:

\vspace{-2mm}

\begin{itemize}
    \item \textbf{CI-Only}: We output only the top cause variable name (e.g., “Occupancy in Zone 3”), without generating a full explanation. Accuracy = 82.0\%, Satisfaction = 2.1.

    
    \item \textbf{FT-Only}: We feed the LLM a random set of three variables (without causal ranking) but fine-tuned as before. Accuracy = 65.3\%, Satisfaction = 3.0.
\end{itemize}

\vspace{-2mm}

These results confirm both components are essential: CI-Only shows the value of causal ranking, but lacks a natural language explanation (hence low satisfaction). FT-Only shows that fine-tuning helps, but without causal priors, cause selection is noisy.

\section{Conclusion}
\label{sec:conlcusion}

We have presented InsightBuild, a two-stage framework that marries explicit causal discovery with LLM-based explanation generation to deliver precise, actionable insights on building energy anomalies. Through experiments on two real-world datasets, InsightBuild outperforms both a rule-based engine and a vanilla LLM by large margins in accuracy and user satisfaction. By ensuring that explanations are rooted in statistically justified causal relations (via Granger tests and structural pruning), our approach mitigates hallucination and provides facility managers with trustworthy reasoning. Future work includes (i) extending to real-time streaming with online causal updates, (ii) integrating domain knowledge (e.g., HVAC thermodynamics) as priors, and (iii) adapting the framework to other cyber-physical systems such as data centers or manufacturing plants.

\nocite{langley00}

\bibliography{example_paper}
\bibliographystyle{icml2025}



\end{document}